\documentclass[10pt,a4paper,reqno]{amsart}

\usepackage{amsfonts}        
\usepackage{graphicx}        
\usepackage{hyperref}        
\usepackage{epsfig}          
\usepackage{anysize}         
\usepackage[dvipsnames]{xcolor} 
\usepackage{wrapfig}         
\usepackage[utf8]{inputenc}  
\usepackage{booktabs}        
\usepackage{textcomp}        

\marginsize{3cm}{3cm}{3cm}{3cm}   
\newcommand{\EASTPaperTitle}{}
\newcommand{\EASTArchive}{}
\newcommand{\EASTAuthors}{}
\newcommand{\EASTAbstract}{}
\newcommand{\EASTKeywords}{}
\newcommand{\EASTAffiliation}{}
\newcommand{\EASTEmail}{}

\newcommand{\PaperTitle}[1]{\gdef\EASTPaperTitle{#1}}
\newcommand{\Archive}[1]{\gdef\EASTArchive{#1}}
\newcommand{\Authors}[1]{\gdef\EASTAuthors{#1}}
\newcommand{\Abstract}[1]{\gdef\EASTAbstract{#1}}
\newcommand{\Keywords}[1]{\gdef\EASTKeywords{#1}}
\newcommand{\Affiliation}[1]{\gdef\EASTAffiliation{#1}}
\newcommand{\Email}[1]{\gdef\EASTEmail{#1}}

\PaperTitle{A Nepali-Accented English Evaluation Dataset for Automatic Speech Recognition}
\Archive{}

\Authors{
    Santosh Dahal \textsuperscript{a},
    Er.\ Kiran Chandra Dahal \textsuperscript{a}
}

\Abstract{
Automatic speech recognition (ASR) systems perform strongly on native-English benchmarks, yet their accuracy degrades sharply when the input speech comes from underrepresented non-native accents. Nepali-accented English is particularly under-served: existing resources either focus on native Nepali speech, cover broader multi-accent settings without dedicated Nepali evaluation, or provide only limited Nepali-accent coverage. This paper presents a Nepali-accented English evaluation dataset designed to support robust ASR benchmarking under accent mismatch. The corpus was collected through a web-based platform that did not collect directly identifying metadata and contains recordings from 57 speakers. Each session follows a fixed 22-prompt protocol consisting of 11 phonetic prompts, 10 domain prompts, and 1 spontaneous prompt, providing complementary coverage of pronunciation, topical vocabulary, and natural speaking style. In addition to transcribed speech, the dataset includes participant metadata for coarse exploratory subgroup analysis and speaker-level manual recording-quality labels. Manual quality assessment shows that 50.9\% of sessions are clean and 42.1\% contain only mild noise. As a descriptive reference, open-source ASR baselines are substantially worse on this corpus than the corresponding LibriSpeech test-clean values reported in official model cards, reaching 38.15--55.00\% WER on the collected set versus reported 2--4\% WER on LibriSpeech test-clean. These baseline results position the corpus as a practical held-out resource for evaluating accent robustness and out-of-distribution generalization on Nepali-accented English.
}

\Keywords{automatic speech recognition, Nepali-accented English, accented speech, evaluation dataset, word error rate}

\Affiliation{
    \textsuperscript{a}
    \textit{Department of Electronics and Computer Engineering, Institute of Engineering, Thapathali Campus, Tribhuvan University}
}

\Email{
    \textsuperscript{a} santosh.080msise16@tcioe.edu.np,
    \textsuperscript{a} dahalkc@tcioe.edu.np
}

\hypersetup{
    pdftitle={\EASTPaperTitle},
    pdfauthor={Santosh Dahal, Kiran Chandra Dahal}
}



\makeatletter
\def\@settitle{%
    \begin{center}%
        \normalfont\LARGE\bfseries\textcolor{NavyBlue}{\@title}%
    \end{center}%
}
\makeatother

\makeatletter
\renewcommand{\section}{%
    \@startsection{section}{1}%
    {\z@}{3.5ex \@plus 1ex \@minus .2ex}{1.15ex\@plus .2ex}%
    {\normalfont\Large\bfseries\color{NavyBlue}}%
}
\makeatother

\usepackage{fancyhdr}
\title{\EASTPaperTitle}
\author{Santosh Dahal\textsuperscript{a}, Kiran Chandra Dahal\textsuperscript{a}}
\date{}

\thanks{
    \textsuperscript{a}\textit{Department of Electronics and Computer Engineering, Institute of Engineering, Thapathali Campus, Tribhuvan University}\\
    \indent E-mail: \href{mailto:santosh.080msise16@tcioe.edu.np}{santosh.080msise16@tcioe.edu.np}; \\
    \indent\hspace*{3.9em}\href{mailto:dahalkc@tcioe.edu.np}{dahalkc@tcioe.edu.np}
}

\begin{document}


    \setcounter{page}{1}
    \vspace{2cm}

    \begin{abstract}
        \EASTAbstract

        \bigskip
        \noindent{\color{NavyBlue}\textbf{Keywords:}} \EASTKeywords
    \end{abstract}

    \maketitle
    \newpage

    \section{Introduction}
    Recent ASR systems achieve very low error rates on standard native-English benchmarks, but these gains do not transfer reliably to accented speech. The gap is especially important for underrepresented accents, where both training and evaluation resources remain scarce. Nepali-accented English belongs to this category: it appears only sparsely in existing corpora even though English is widely used in education, technology, public services, and professional communication in Nepal.

    The lack of a focused benchmark makes it difficult to answer a basic question: how well do current ASR models handle English spoken by Nepali L1 speakers? Multi-accent resources such as L2-ARCTIC provide valuable non-native English data, but they do not include Nepali speakers \cite{Zhao2018L2ARCTIC}. Other accented-English corpora cover a broader set of foreign accents yet still omit Nepali as a target evaluation accent \cite{Lander2022CSLU}. Meanwhile, Svarah demonstrates the importance of evaluating ASR on Indian accents and includes a Nepali-accented English category, but the speakers in that subset are from India rather than Nepal. Its scope is also broader than a Nepali-specific benchmark and is not centered on controlled phonetic coverage for this accent \cite{javed2023svarahevaluatingenglishasr}. Existing Nepali ASR work, in contrast, focuses on speech recognition for the Nepali language itself rather than English spoken with a Nepali accent \cite{bhatta2020nepali,banjara2020nepali,dhakal2022automatic}.

    This paper presents a dataset created to fill that gap. The corpus is designed as an \emph{evaluation-focused} benchmark rather than a large-scale pretraining resource. The collection protocol aims to keep per-speaker coverage comparable while still capturing three complementary dimensions of variation: phonetic coverage, domain vocabulary, and spontaneous speech. The resulting dataset is then used as a held-out benchmark for ASR evaluation.

    The main contributions of this paper are as follows:

    \begin{itemize}
        \item A Nepali-accented English evaluation corpus with 57 speakers collected through a consistent prompt-based protocol.
        \item A prompt design that combines phonetic prompts, domain prompts, and spontaneous speech within every participant session.
        \item Metadata and speaker-level recording-quality annotations that support coarse exploratory subgroup analysis and data-quality filtering.
        \item Baseline evidence that modern open-source ASR systems degrade substantially on the corpus compared with their reported LibriSpeech performance.
    \end{itemize}

    \section{Related Work}

    Several existing datasets motivate the need for a Nepali-specific accented-English benchmark. L2-ARCTIC is a widely used non-native English corpus containing six L1 accent groups and carefully read speech, making it valuable for accent adaptation research; however, it does not include Nepali-accented English \cite{Zhao2018L2ARCTIC}. The CSLU Foreign Accented English corpus broadens accent coverage, but it is likewise not targeted toward Nepali speakers \cite{Lander2022CSLU}. Svarah extends the evaluation landscape by covering Indian-accented English and explicitly includes a Nepali-accented speech subset, showing that ASR systems experience pronounced degradation when moving away from native-English benchmarks. However, the speakers in that subset are from India rather than Nepal \cite{javed2023svarahevaluatingenglishasr}. Still, Svarah is a multilingual-accent benchmark spanning many language backgrounds and speaking conditions; it does not offer a Nepali-only evaluation design with fixed per-speaker phonetic and topical coverage.

    On the Nepali side, prior work has concentrated on ASR for native Nepali speech using convolutional, recurrent, and Conformer-style architectures \cite{bhatta2020nepali,banjara2020nepali,dhakal2022automatic}. These studies are important for low-resource speech technology in Nepal, but they address a different problem from recognizing English produced with Nepali phonetic transfer. The present corpus therefore complements both lines of work: it is not a replacement for broad multi-accent English datasets, and it is not a native Nepali ASR corpus; instead, it specifically targets evaluation of Nepali-accented English.

    \section{Dataset Design and Collection Protocol}

    The dataset was built around a simple design principle: each participant should contribute a comparable but diverse set of utterances so that the final corpus supports controlled evaluation rather than ad hoc sampling. Table~\ref{tab:dataset_overview} summarizes the overall design.

    \begin{table}[t]
    \centering
    \caption{Overview of the Nepali-accented English evaluation dataset.}
    \label{tab:dataset_overview}
    \small
    \begin{tabular}{@{}p{0.30\linewidth}p{0.62\linewidth}@{}}
    \toprule
    \textbf{Aspect} & \textbf{Description} \\
    \midrule
    Primary goal & Held-out evaluation benchmark for Nepali-accented English ASR \\
    Participants & 57 speakers \\
    Collection mode & Remote recording through a web-based platform without directly identifying metadata \\
    Speech styles & Controlled read speech and one spontaneous response per session \\
    Session structure & 22 prompts per participant \\
    Prompt mix & 11 phonetic prompts, 10 domain prompts, 1 spontaneous prompt \\
    Metadata & Age range, gender, highest qualification, occupation, ethnicity, district of origin, and topic of interest \\
    Quality annotation & Manual speaker-level recording-quality label on a five-point ordinal scale \\
    Intended use & Accent-robust ASR evaluation, coarse subgroup analysis, and error analysis \\
    \bottomrule
    \end{tabular}
    \end{table}

    \subsection{Collection Platform and Workflow}

    Data were collected through a dedicated single-page web application designed for guided remote participation. The participant workflow contains four stages: session initialization without directly identifying metadata, consent acknowledgement with metadata entry, prompt-wise recording with re-record or retry options, and final review with submission. This structure was chosen to keep the data collection experience lightweight while still improving consistency across speakers and sessions.

    The platform records only analysis-oriented metadata needed for curation and descriptive analysis. Session information is stored together with timestamps and completion status, allowing the final evaluation set to be curated without collecting direct personal identifiers. Because speech recordings can still be identifying, any dataset distribution or reuse should remain within participant-consent boundaries and appropriate access control.

    \subsection{Prompt Design}

    Each participant session contains 22 prompts drawn from three prompt types, shown in Table~\ref{tab:prompt_types}. The phonetic prompts target pronunciation coverage, the domain prompts broaden lexical and topical variation, and the spontaneous prompt captures less constrained speech with more natural discourse structure.

    \begin{table}[t]
    \centering
    \caption{Prompt design used in each recording session.}
    \label{tab:prompt_types}
    \small
    \begin{tabular}{lcp{0.48\linewidth}}
    \toprule
    \textbf{Prompt type} & \textbf{Count} & \textbf{Purpose} \\
    \midrule
    Phonetic & 11 & Cover English phonemes and common phonotactic contexts using CMU Arctic-style prompts \\
    Domain & 10 & Introduce topic-specific vocabulary from domains such as technology, science, history, environment, culture, health, education, geography, and society \\
    Spontaneous & 1 & Elicit natural speech from a participant-selected interest topic such as sports, reading, music, or cooking \\
    \bottomrule
    \end{tabular}
    \end{table}

    The phonetic prompts were designed in the style of the CMU Arctic prompt inventory \cite{kominek2003cmuarctic} and organized to keep phonetic coverage balanced across participants. Domain prompts were curated from public text sources, then normalized for readability and collection consistency. The spontaneous prompt is tied to the participant's self-reported topic of interest so that the dataset includes at least one segment with less scripted speaking behavior per session.

    \subsection{Transcription of Spontaneous Speech}

    Spontaneous responses were derived from elicited recordings whose metadata designated the spontaneous condition, as distinct from read phonetic or domain-matched material. For each item, the corresponding waveform was assigned a short speaker identifier and a unique session--utterance ID, and stored under the \texttt{spontaneous} partition of the corpus.

    \paragraph{Segmentation.}
    Recordings longer than 20~s were split prior to automatic transcription. Silence boundaries were approximated with \textsc{ffmpeg} \texttt{silencedetect} using a noise threshold of $-35$~dB and a minimum silence duration of $1.0$~s. Resulting speech intervals were further constrained so that no segment exceeded 20~s, with a minimum segment length of $0.25$~s. Recordings of 20~s or less were left as a single segment.

    \paragraph{Automatic speech recognition.}
    Each segment was transcribed with the Google Generative Language API using the Python \texttt{google-genai} client. The generative model was \texttt{gemini-3.1-pro-preview}. For each call, the model received the audio in \textsc{wav} form (\texttt{audio/wav}) together with a fixed text instruction. Whitespace in the model output was normalized; an empty hypothesis was treated as a no-speech case. Table~\ref{tab:asr} summarizes the ASR back end used for spontaneous transcription.

    \paragraph{Instruction prompt (exact text).}
    The following prompt was used for transcription:
    \begin{quote}
    \raggedright\small\ttfamily
    Listen to this audio and transcribe the speech.\\
    Rules:\\
    - Output only the spoken words, in English (including any Nepali-accented English as heard).\\
    - No timestamps, no speaker labels, no ``Transcript:'' prefix, no markdown.\\
    - If there is no intelligible speech, output exactly: [NO\_SPEECH]
    \end{quote}

    \paragraph{Transcript assembly.}
    For multi-segment recordings, per-segment transcriptions were concatenated in temporal order with inter-word spaces to form one line per session--utterance, stored in \texttt{trans.txt} in the form \textit{utterance-id} followed by the full transcript, consistent with the read side of the corpus. Segments for which the model returned the token \texttt{[NO\_SPEECH]} (exactly) were treated as non-contributing at the segment level when assembling the line.

    \begin{table}[t]
    \centering
    \caption{ASR back end used for spontaneous transcription.}
    \label{tab:asr}
    \small
    \begin{tabular}{@{}p{0.22\linewidth}p{0.62\linewidth}@{}}
    \toprule
    Item & Value \\
    \midrule
    Model & \texttt{gemini-3.1-pro-preview} \\
    No-speech token & \texttt{[NO\_SPEECH]} (exact) \\
    \bottomrule
    \end{tabular}
    \end{table}

    \subsection{Metadata and Privacy}

    The collection form supports age range (18--30, 30--45, 45--60, 60+), gender, highest qualification, occupation domain, ethnicity category, district of origin, and topic of interest. In the present paper, we report six summary attributes for analysis: gender, ethnicity, age group, highest qualification, occupation domain, and topic of interest. District of origin was collected during data acquisition but is not broken down in the aggregate tables below.

    \section{Corpus Statistics and Recording Quality}

\begin{figure}[t]
    \centering
    \includegraphics[width=0.82\linewidth]{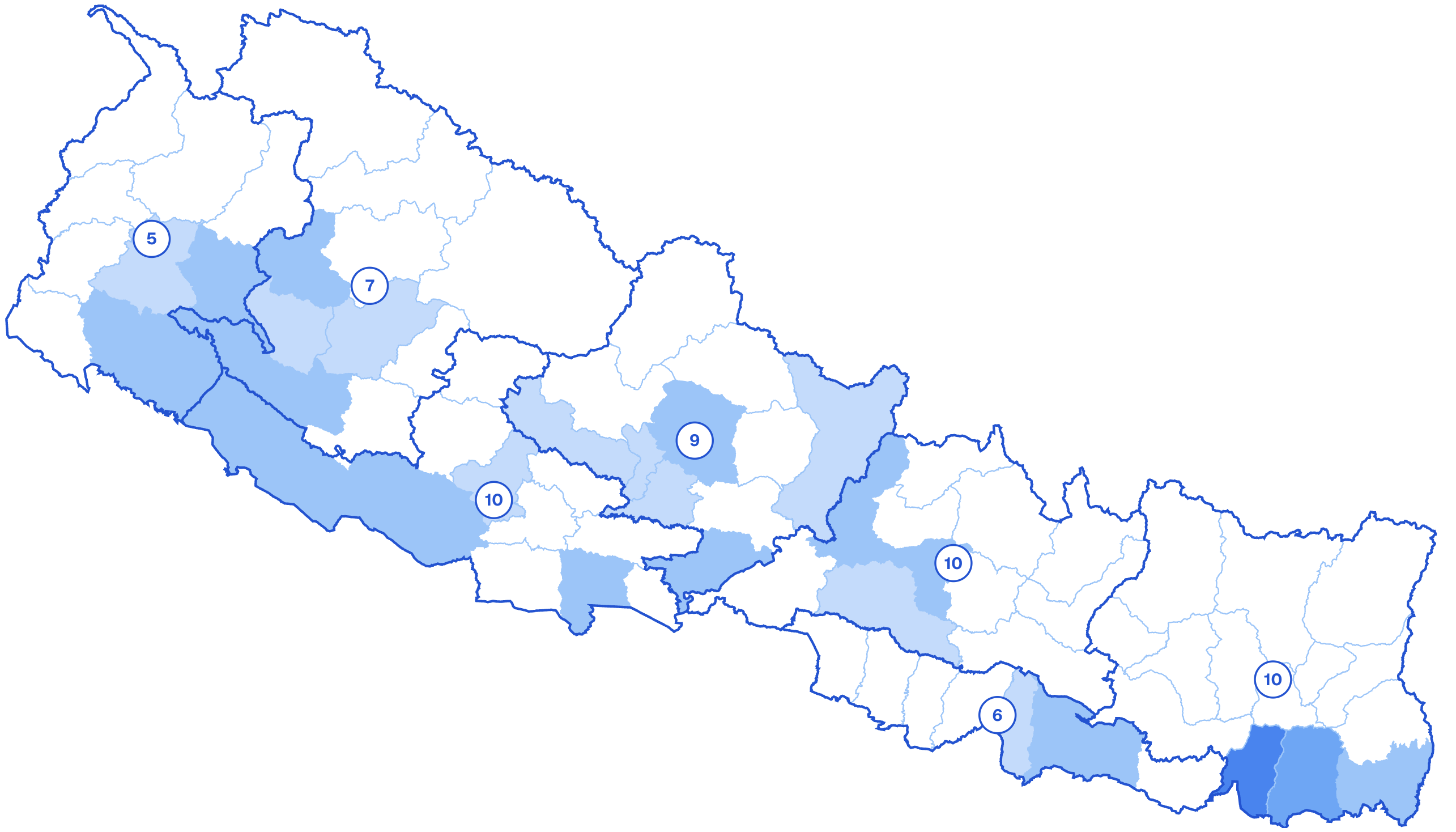}
    \caption{Geographic distribution of participants by reported district of origin across Nepal.}
    \label{fig:participant_map}
    \end{figure}

    The final evaluation set contains 57 participants. Figure~\ref{fig:participant_map} shows the geographic distribution of participants by reported district of origin across Nepal, while Tables~\ref{tab:participant_profile} and \ref{tab:participant_context} summarize the collected metadata distributions.

    \begin{table}[t]
    \centering
    \caption{Participant profile summary (\(n=57\)).}
    \label{tab:participant_profile}
    \small
    \begin{tabular}{llrr}
    \toprule
    \textbf{Attribute} & \textbf{Category} & \textbf{Count} & \textbf{\%} \\
    \midrule
    Gender & Male & 37 & 64.9 \\
    Gender & Female & 20 & 35.1 \\
    \addlinespace
    Age group & 18--30 & 45 & 78.9 \\
    Age group & 30--45 & 10 & 17.5 \\
    Age group & 45--60 & 2 & 3.5 \\
    \addlinespace
    Qualification & Bachelors & 38 & 66.7 \\
    Qualification & Masters & 12 & 21.1 \\
    Qualification & +2 (higher secondary) & 7 & 12.3 \\
    \bottomrule
    \end{tabular}
    \end{table}

    \begin{table}[t]
    \centering
    \caption{Participant context attributes (\(n=57\)).}
    \label{tab:participant_context}
    \small
    \begin{tabular}{llrr}
    \toprule
    \textbf{Attribute} & \textbf{Category} & \textbf{Count} & \textbf{\%} \\
    \midrule
    Ethnicity & Brahmin/Chhetri & 43 & 75.4 \\
    Ethnicity & Madhesi & 6 & 10.5 \\
    Ethnicity & Janajati & 4 & 7.0 \\
    Ethnicity & Other & 2 & 3.5 \\
    Ethnicity & Newar & 1 & 1.8 \\
    Ethnicity & Dalit & 1 & 1.8 \\
    \addlinespace
    Occupation & Engineering/Technology & 31 & 54.4 \\
    Occupation & Student & 13 & 22.8 \\
    Occupation & Health/Medicine & 3 & 5.3 \\
    Occupation & Business/Management & 3 & 5.3 \\
    Occupation & Education/Teaching & 2 & 3.5 \\
    Occupation & Science & 2 & 3.5 \\
    Occupation & Law/Government & 2 & 3.5 \\
    Occupation & Other & 1 & 1.8 \\
    \addlinespace
    Interest & Travelling & 17 & 29.8 \\
    Interest & Sports & 14 & 24.6 \\
    Interest & Reading & 9 & 15.8 \\
    Interest & Movies/Entertainment & 5 & 8.8 \\
    Interest & Painting/Drawing & 4 & 7.0 \\
    Interest & Music & 4 & 7.0 \\
    Interest & Cooking & 3 & 5.3 \\
    Interest & Stitching/Sewing & 1 & 1.8 \\
    \bottomrule
    \end{tabular}
    \end{table}

    The collected subset is not demographically balanced, but it does capture meaningful variation in age, education, ethnicity, occupation, and self-selected topics. The distribution is dominated by younger adults (78.9\% in the 18--30 range), participants with bachelor's-level education (66.7\%), and contributors from engineering or technology backgrounds (54.4\%). Topic-of-interest responses are led by travelling, sports, and reading, which helps diversify the spontaneous-speech portion of the corpus.

    \subsection{Speaker-Level Recording Quality}

    Because all utterances from a participant were recorded within a single session, ambient noise, microphone placement, and room acoustics are largely session-constant. For that reason, recording quality was summarized at the speaker level rather than the utterance level. The annotation procedure sampled three utterances uniformly at random for each speaker, played them under fixed listening conditions, and assigned one of five ordinal labels: clean, mild noise, noisy, very noisy, or bad recording.

    Table~\ref{tab:recording_quality} shows the resulting distribution. The large majority of sessions (93.0\%) fall in the clean or mild-noise categories, suggesting that the dataset is generally suitable for ASR benchmarking while still preserving realistic variability from remote volunteer recordings. These labels should be interpreted as coarse speaker-level tags for exploratory filtering and error analysis rather than as a formally validated acoustic annotation layer.

    \begin{table}[t]
    \centering
    \caption{Manual speaker-level recording-quality labels.}
    \label{tab:recording_quality}
    \small
    \begin{tabular}{lrr}
    \toprule
    \textbf{Label} & \textbf{Count} & \textbf{\%} \\
    \midrule
    Clean & 29 & 50.9 \\
    Mild noise & 24 & 42.1 \\
    Noisy & 3 & 5.3 \\
    Very noisy & 0 & 0.0 \\
    Bad recording & 1 & 1.8 \\
    \bottomrule
    \end{tabular}
    \end{table}

    \section{Benchmark Evaluation with Open-Source ASR Models}

    To assess the difficulty of the corpus, the dataset was used as a held-out benchmark for several open-source ASR models. Table~\ref{tab:opensource_baselines} contrasts the combined WER on the collected set with each model's reported LibriSpeech test-clean WER from the corresponding official model card \cite{nvidia2024conformerctcsmall,nvidia2024conformerctclarge,openai2022whispersmall,openai2022whisperlarge}. Because the LibriSpeech values are taken from official model cards rather than a single re-run pipeline, the comparison should be interpreted as a descriptive cross-dataset reference rather than a controlled estimate of accent-only degradation. For fully reproducible reuse, the benchmark release should package the corresponding reference transcripts, normalization rules, and scoring scripts alongside the audio and metadata.

    \begin{table}[t]
    \centering
    \caption{Reported LibriSpeech test-clean WER from official model cards and observed WER on the collected Nepali-accented English set.}
    \label{tab:opensource_baselines}
    \small
    \begin{tabular}{lcc}
    \toprule
    \textbf{Model} & \textbf{LibriSpeech test-clean} & \textbf{Collected set} \\
    \midrule
    Conformer Small & 3.70 & 46.00 \\
    Conformer Large & 2.20 & 38.30 \\
    Whisper Small & 3.43 & 55.00 \\
    Whisper Large & 3.00 & 38.15 \\
    \bottomrule
    \end{tabular}
    \end{table}

    All tested models degrade substantially on the collected benchmark. Even the strongest large-model baselines, Conformer Large and Whisper Large, rise from approximately 2--3\% WER on LibriSpeech test-clean to about 38\% WER on the Nepali-accented set. The smaller models degrade even more sharply, with WER reaching 46.00\% for Conformer Small and 55.00\% for Whisper Small. These descriptive results reinforce the dataset's usefulness as a stress test for accent robustness and show that strong native-English benchmark performance does not necessarily imply reliable performance on Nepali-accented English under the present evaluation setup.

    \section{Discussion, Intended Use, and Limitations}

    The dataset has three main strengths. First, it combines controlled phonetic coverage with broader topical prompts and spontaneous speech, making it more informative than a read-only benchmark. Second, the metadata support coarse exploratory subgroup and robustness analysis beyond a single aggregate WER number, although several categories remain too small for strong comparative claims. Third, the recording-quality labels make it possible to analyze or filter the effect of session noise on ASR performance.

    At the same time, the current corpus has clear limitations. The participant pool is skewed toward younger, highly educated speakers and is dominated by engineering or technology backgrounds. The recordings are also collected remotely rather than under studio conditions, which introduces device and environment variability even when most sessions remain usable. Finally, the dataset is best suited to held-out evaluation and error analysis; its size and collection design are not intended to replace broader multi-accent training corpora.

    We therefore recommend the corpus for three use cases in particular. The first is benchmarking accent-robust ASR. The second is analyzing errors across prompt types or coarse participant groupings. The third is measuring out-of-distribution generalization when Nepali-accented English is absent from training data. We do not recommend using the metadata for speaker profiling or any ranking of participants; the metadata are included to support coarse exploratory subgroup and robustness analysis, not identity inference.

    \section{Conclusion}

    This paper presented a Nepali-accented English evaluation dataset for ASR containing recordings from 57 speakers collected through a structured web-based protocol. The dataset pairs fixed per-session prompt coverage with demographic and quality annotations, creating a practical benchmark for studying accent robustness. Baseline experiments with open-source Conformer and Whisper models show substantially higher WER on the collected set than the corresponding published LibriSpeech reference values, reinforcing the need for targeted evaluation on underrepresented accents. The corpus provides a useful foundation for future work on Nepali-accented English recognition, accent-aware adaptation, and more inclusive ASR evaluation.

    The dataset is publicly available on GitHub at \href{https://github.com/santoshdahal2016/Nepali-Accented-English-Evaluation-Dataset}{the project repository}.

    \bibliographystyle{reference}
    \bibliography{reference}

@misc{javed2023svarahevaluatingenglishasr,
  author = {Tahir Javed and Sakshi Joshi and Vignesh Nagarajan and Sai Sundaresan and Janki Nawale and Abhigyan Raman and Kaushal Bhogale and Pratyush Kumar and Mitesh M. Khapra},
  title = {Svarah: Evaluating {English} {ASR} Systems on {Indian} Accents},
  year = {2023},
  eprint = {2305.15760},
  archiveprefix = {arXiv},
  primaryclass = {cs.CL},
  note = {arXiv:2305.15760 [cs.CL]},
  url = {https://arxiv.org/abs/2305.15760}
}

@inproceedings{bhatta2020nepali,
  author = {B. Bhatta and B. Joshi and R. K. Maharjan},
  title = {{Nepali} Speech Recognition Using {CNN}, {GRU} and {CTC}},
  booktitle = {Proceedings of the 32nd Conference on Computational Linguistics and Speech Processing (ROCLING 2020)},
  pages = {238--246},
  year = {2020}
}

@inproceedings{banjara2020nepali,
  author = {J. Banjara and K. R. Mishra and J. Rathi and K. Karki and S. Shakya},
  title = {{Nepali} Speech Recognition Using {CNN} and Sequence Models},
  booktitle = {2020 IEEE International Conference on Machine Learning and Applied Network Technologies (ICMLANT)},
  pages = {1--5},
  year = {2020},
  publisher = {IEEE}
}

@inproceedings{dhakal2022automatic,
  author = {M. Dhakal and A. Chhetri and A. K. Gupta and P. Lamichhane and S. Pandey and S. Shakya},
  title = {Automatic Speech Recognition for the {Nepali} Language Using {CNN}, Bidirectional {LSTM} and {ResNet}},
  booktitle = {2022 International Conference on Inventive Computation Technologies (ICICT)},
  pages = {515--521},
  year = {2022},
  publisher = {IEEE}
}

@techreport{kominek2003cmuarctic,
  author = {John Kominek and Alan W. Black},
  title = {The {CMU ARCTIC} Speech Databases},
  institution = {Language Technologies Institute, Carnegie Mellon University},
  year = {2003},
  number = {CMU-LTI-03-177}
}

@misc{Lander2022CSLU,
  author = {Lander, T.},
  title = {{CSLU}: Foreign Accented {English} Release 1.2},
  year = {2022},
  note = {Dataset, Release 1.2},
  doi = {10.5683/SP2/K7EQTE},
  url = {https://doi.org/10.5683/SP2/K7EQTE}
}

@inproceedings{Zhao2018L2ARCTIC,
  author = {Guoguo Zhao and Sertaç Sonsaat and Anan Silpachai and Ivan Lucic and Ekaterina Chukharev-Hudilainen and John Levis and Ricardo Gutierrez-Osuna},
  title = {{L2-ARCTIC}: A Non-Native {English} Speech Corpus},
  booktitle = {Interspeech 2018},
  pages = {2783--2787},
  year = {2018},
  doi = {10.21437/Interspeech.2018-1110}
}

@misc{nvidia2024conformerctcsmall,
  author = {{NVIDIA}},
  title = {{NVIDIA Conformer-CTC Small} ({en-US})},
  year = {2024},
  note = {{Hugging Face} model card, accessed 2026-04-24},
  url = {https://huggingface.co/nvidia/stt_en_conformer_ctc_small}
}

@misc{nvidia2024conformerctclarge,
  author = {{NVIDIA}},
  title = {{NVIDIA Conformer-CTC Large} ({en-US})},
  year = {2024},
  note = {{Hugging Face} model card, accessed 2026-04-24},
  url = {https://huggingface.co/nvidia/stt_en_conformer_ctc_large}
}

@misc{openai2022whispersmall,
  author = {{OpenAI}},
  title = {{Whisper Small}},
  year = {2022},
  note = {{Hugging Face} model card, accessed 2026-04-24},
  url = {https://huggingface.co/openai/whisper-small}
}

@misc{openai2022whisperlarge,
  author = {{OpenAI}},
  title = {{Whisper Large}},
  year = {2022},
  note = {{Hugging Face} model card, accessed 2026-04-24},
  url = {https://huggingface.co/openai/whisper-large}
}


\end{document}